\documentclass[10pt,twocolumn,letterpaper]{article}

\usepackage{cvpr}
\usepackage{times}
\usepackage{epsfig}
\usepackage{graphicx}
\usepackage{amsmath}
\usepackage{amssymb}
\usepackage{array}
\usepackage{bm}
\usepackage{booktabs}
\usepackage{caption3}
\usepackage{cases}
\usepackage{indentfirst}
\usepackage{mathrsfs}
\usepackage{multirow}
\usepackage{url}
\usepackage{xcolor}
\usepackage{algorithm}
\usepackage{algorithmic}

\usepackage[breaklinks=true,bookmarks=false]{hyperref}

\cvprfinalcopy 

\def\cvprPaperID{****} 

\begin{document}

\title{Robust Visual Tracking via Implicit Low-Rank Constraints and Structural Color Histograms}

\author{Yi-Xuan Wang$^{1}$, Xiao-Jun Wu$^{1, *}$, Xue-Feng Zhu$^{1}$\\
$^1$School of Internet of Things Engineering, Jiangnan University, Wuxi, China.\\
{\tt\small xiaojun\_wu\_jnu@163.com}
}

\maketitle

\begin{abstract}
With the guaranteed discrimination and efficiency of spatial appearance model, Discriminative Correlation Filters (DCF-) based tracking methods have achieved outstanding performance recently. 
However, the construction of effective temporal appearance model is still challenging on account of filter degeneration becomes a significant factor that causes tracking failures in the DCF framework. 
To encourage temporal continuity and to explore the smooth variation of target appearance, we propose to enhance low-rank structure of the learned filters, which can be realized by constraining the successive filters within a $\ell_2$-norm ball. 
Moreover, we design a global descriptor, structural color histograms, to provide complementary support to the final response map, improving the stability and robustness to the DCF framework. 
The experimental results on standard benchmarks demonstrate that our Implicit Low-Rank Constraints and Structural Color Histograms (ILRCSCH) tracker outperforms state-of-the-art methods.
\end{abstract}

\section{Introduction}

Object tracking is one of the most fundamental problems in computer vision due to its extensive and numerous applications in practical scenarios, such as human-computer interaction, video surveillance and robotics. 
The main objective of visual object tracking is to track a specific object and estimate its trajectory by initializing the object of interest in the first frame with a bounding box. 
Despite the significant progress in recent years, visual tracking still remains challenging to design a robust tracker due to factors such as shape deformation, illumination variations, background clutter and scale changes.

There are two main approaches to cope with the issues mentioned above, namely generative and discriminative methods. 
The generative methods address the tracking problem by searching for the regions that most similar to the target appearance. 
These methods are usually based on template matching~\cite{Adam2006Robust,Comaniciu2003Kernel} or subspace learning~\cite{Black1996EigenTracking,Ross2008Incremental}. 
Recently, sparse linear representation methods combining with the particle filter framework via solving $\ell_1$-norm minimization problem have been proposed for object tracking~\cite{Mei2009Robust}. 
Because of the short interval, there is only a slight change of the target appearance and corresponding surroundings between successive frames.
The vectorized frames within a short a period time should be linearly correlated. 
Therefore, the methods with low-rank constraints~\cite{Zhang2015Robust} have showed their  robustness in dealing with image corruptions caused by partial occlusions or illumination variations. 
However, there are several defects of these methods. 
First, the formulations of solving sparse and low-rank problems are time-consuming.
Second, the lack of training samples entails the dictionary in sparse representation not reliable.

\begin{figure}[tbp]
\begin{center}
   \includegraphics[width=1\linewidth]{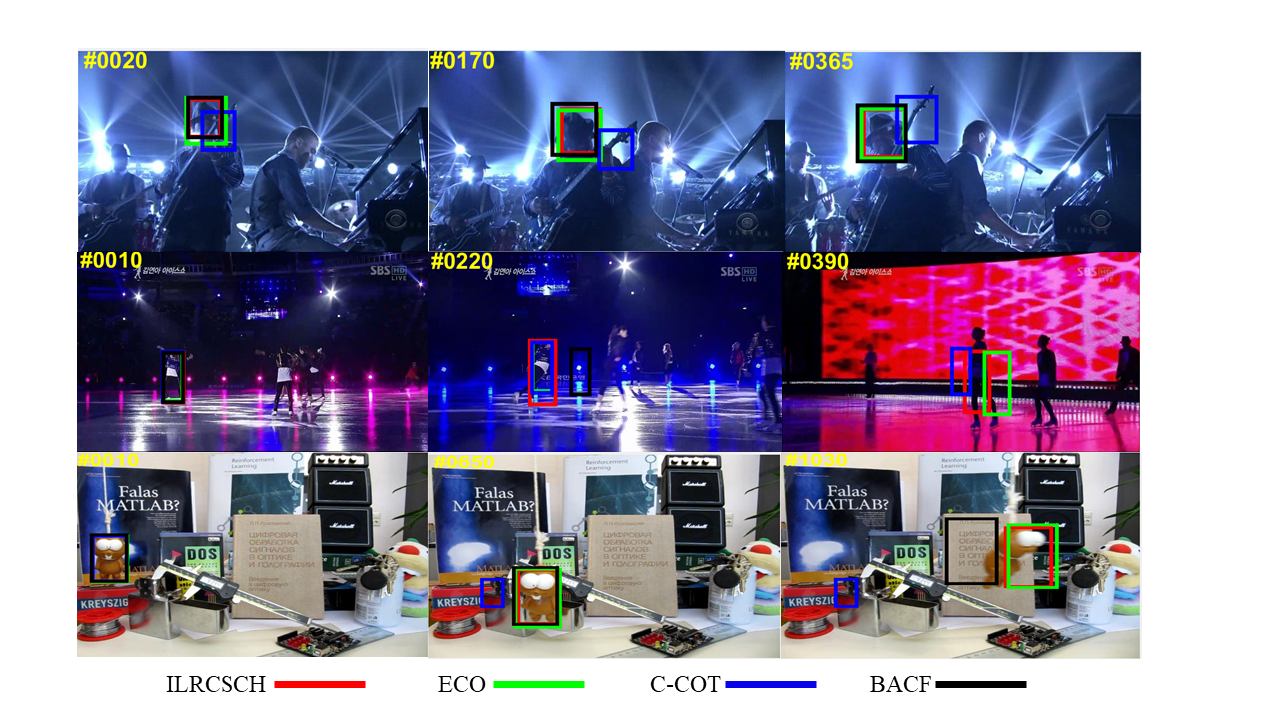}   
\end{center}
   \caption{Comparisons of the proposed ILRCSCH tracker with the
state-of-the-art correlation filter trackers (ECO~\cite{Danelljan2016ECO}, CCOT~\cite{Danelljan2016Beyond}, and BACF~\cite{Galoogahi2017Learning}) on the \textit{Shaking}, \textit{Skating1}, and \textit{Lemming} sequences~\cite{Wu2015Object}.}\label{fig1}
\end{figure}

To this end, approaches based on Discriminative Correlation Filter (DCF)~\cite{Bolme2010Visual} have been proposed and shown great performance on many tracking benchmarks \cite{Kristan2016The, kristan2017visual, kristan2018sixth, kristan2019seventh, Wu2015Object} in recent years. 
The core component of a standard DCF tracker is to train a classifier to distinguish between the target and its background. 
By exploiting Fast Fourier Transform (FFT) for all training samples generated by cyclic shift operations at the training and detection steps, this type of methods have very low computational complexity and enable  over hundreds of frames-per-second. 
DCF tracker with handcrafted features HOG (KCF)~\cite{Henriques2015High,feng2015random} proposed by Henriques presents great performance and high efficiency in most general tracking cases. 
It should be noted that the speed of KCF~\cite{Henriques2015High} tracker is among the fastest ones of all tracking algorithms currently.
Nevertheless, KCF~\cite{Henriques2015High} has its limitations. During the tracking process, as the object's appearance can vary significantly, the learned template model based on HOG features are not good at handling rotations and deformations. 
When the model templates need to be updated, the small errors could be accumulated, which may eventually cause model drift or even tracking failure. 
In order to alleviate the insufficiency of appearance representation, more recent DCF trackers use a discriminative object representation with multi-handcrafted features such as HOG in combine with Color Names (CN)~\cite{Weijer2009Learning, xu2018non}. 
Latest work has also integrated deep Convolutional Neural Networks into the DCF tracking framework \cite{Danelljan2016ECO, xu2019Joint, wang2018multi}. 
The parameters of the network are pre-trained on a large amount of labeled image datasets, \textit{e.g.}, the ImageNet~\cite{Russakovsky2015ImageNet}. 
Deep convolutional features obtained by such pre-trained neural network have been demonstrated to be more discriminative and robust~\cite{Nam2015Learning, xu2019learning} than traditional handcrafted features. 
However, either extracting or applying these features is to be quite computationally expensive. 

In spite of many efforts and significant improvements in recent years, object tracking still remains challenging for real-world applications. 
In this work, we propose Implicit Low-Rank Constraints and Structural Color Histograms (ILRCSCH), which is based on three handcrafted features (HOG, CN~\cite{Weijer2009Learning}, and Structural Color Histogram) in order to make realize adaptive trade-off between accuracy and efficiency in visual tracking. 
Our approach takes the history information into account during learning process, the vectorized filters obtained by our tracking framework tend to be low-rank. 
Therefore, they can be added in a low dimensional subspace as the result of template model history information. 
Furthermore, in order to suppress the boundary effects, our approach puts mask on the learned filter in the first frame. 
The mask can also be considered as a low-rank reference information due to that there is no template model for the first frame in visual tracking. 
The experimental results on several recent standard datasets demonstrate that our approach outperforms state-of-art DCF-based trackers in terms of accuracy and robustness without sacrificing efficiency (qualitative comparisons are illustrated in Fig.~\ref{fig1}).

In the reminder of the paper, we firstly present a brief introduction of techniques that most related to our work in Section~\ref{related_work}. 
In Section~\ref{tracker}, we introduce the overview and details about the proposed ILRCSCH tracking framework. 
The experimental results and discussions are given in Section~\ref{experiment}. 
In Section~\ref{conclusion}, we make conclusions of this work.

\section{Related work}\label{related_work}

Many efforts have been made in the field of object tracking research. In this section, we briefly present the most relevant approaches to our work. A comprehensive review is recommended in recent benchmarks~\cite{Wu2013Online,Wu2015Object,Kristan2016The,Liang2015Encoding}.

\textbf{DCF Tracking Framework}: In recent years, DCF tackers have gained extensive attention, due to their high accuracy and efficiency. 
DCF trackers usually follow the tracking by detection paradigm, which was first proposed in TLD~\cite{Kalal2012Tracking}. 
Typical DCF trackers learn and update a correlation filter online to detect the object target in consecutive frames. 
In the early work, DCF trackers were restricted to a single feature channel. 
Bolme \textit{et al.} initially proposed Minimum Output Sum Of Squared Error (MOSSE)~\cite{Bolme2010Visual} for visual tracking on gray-scale images. 
By using correlation filters, MOSSE is able to run over hundreds of frames per second. 
Herques \textit{et al.}  exploited circular shift to learn correlation filters in a kernel space (CSK)~\cite{Henriques2012Exploiting}, which runs at the highest speed among recent benchmarks. 
This work was later extended to multi-channel feature maps by HOG (KCF)~\cite{Henriques2015High} and color features (CN)~\cite{Danelljan2014Adaptive}. 
Moreover, a remarkable improvement was achieved by DSST~\cite{danelljan2014accurate} tracker which learns adaptive multi-scale correlation filters with HOG features. 
Zhang \textit{et al.} also proposed spatial-context information (STC)~\cite{Zhang2014Fast} which incorporated Bayesian framework into filter learning process to cope with the partial occlusion of object targets. 
However, in real applications for tracking, the accuracy of STC heavily relies on the selection of the parameter values. 
Despite their success, almost all the typical DCF-based trackers suffer from the boundary effects due to the circulant assumption. 
A typical DCF-based tracker usually utilizes the circulant assumption of a local image patch to perform efficient training and detection by using Fast Fourier Transform (FFT). 
However, this operation brings the unwanted boundary effects, which restricts the size of the image region used for training and detection. 
The boundary effects are often suppressed by a cosine window.
Unfortunately, the cosine window will again heavily reduce the search region and lead to degrade the tracker's ability to distinguish target from its background.
In recent years, several methods were proposed to solve this issue, Danelljan \textit{et al.} proposed Spatially Regularised Discriminative Correlation Filter (SRDCF)~\cite{Danelljan2015Learning} via adding a spatial regularization term into the objective function. 
In order to enhance the discriminability of the filter, Danelljan \textit{et al.} also utilized convolutional features which are obtained from a deep RGB network on SRDCF tracker (DeepSRDCF)~\cite{Danelljan2016Convolutional}. 
Unfortunately, this was achieved at the cost of high computational complexity. 
Moreover, some parameters in SRDCF tracker must be carefully tuned, otherwise it can lead to poor tracking performance. 
Similar to SRDCF, Galoogahi proposed Learning Background-Aware Correlation filters for Visual Tracking (BACF)~\cite{Galoogahi2017Learning}. 
Unlike the traditional DCF trackers where negative samples are limited to circular shifted patches, BACF tracker is learning and updating filters from real negative samples which densely cropped from the background area by using a binary matrix. 
Lukezic addressed the problem by proposing spatial reliability map (CSRDCF)~\cite{Lukezic2017Discriminative}. 
In CSRDCF, the filter is equipped with a color-based binary mask when the discriminative target region is activated. 
The limitation of CSRDCF is that the mask for the training patches over rely on the color features. Particularly, in the practical tracking process, the robustness of color features is relatively poor comparing to HOG. 
In order to make a trade-off between accuracy and efficiency, Bertinetto proposed Sum of Template and Pixel-wise Learners (STAPLE)~\cite{Bertinetto2016Staple} which learns multiple handcrafted features with HOG and color histogram. 
It is well known that HOG features are very sensitive to deformation but color statistics can cope well with variation in shape. 
Therefore, STAPLE uses the combination to cope with variations of object's appearance. 
Instead of fusing the predictions of multiple models, STAPLE  combines the scores of correlation filter models and color-based models at the final step to improve robustness. 
Recently this work was improved by Context-Aware Correlation Filter-Tracking (Staple CA)~\cite{mueller2017context} which is quite similar to the work of BACF~\cite{Galoogahi2017Learning}. Staple CA~\cite{mueller2017context} takes in the context information into account, and learns from background patches as negative samples for filter training. 
Danelljan \textit{et al.} proposed C-COT~\cite{Danelljan2016Beyond} tracker which introduced a continuous-domain formulation of the DCF paradigm. 
C-COT achieved top performance in the VOT2016~\cite{Kristan2016The} by enabling DCF tracker with multi-resolution deep features.
However, this was built by sacrificing the real-time capabilities, suppressing the advantage of the early DCF-based trackers.
The subsequent tacker ECO~\cite{Danelljan2016ECO} further optimized C-COT in terms of efficiency. 
Then, group feature selection is emphasized in LADCF~\cite{xu2018learning} to realize adaptive spatial configuration.
In VOT2017, half of the top 10 trackers were based either ECO or its predecessor C-COT. 
In this paper, we proposed ILRCSCH tracker by exploiting $ \ell_2$-norm and three handcrafted features, and we extensively evaluated our method with the state-of-art trackers which contains ECO and C-COT on several recent standard benchmarks with superior results, and the speed of our tracker achieves nearly three times faster than C-COT.

\textbf{Sparse Linear Representation Tracking Framework}: Tracking approaches based on Sparse linear representation have proved to be simple yet effective among generative tracking methods. 
The most important part of this type of approaches is that how to construct and optimize the target appearance model. 
Mei and Ling~\cite{Mei2009Robust} proposed the seminal tracking method based on solving $\ell_1$ minimization problem. 
By using the sparsity constraint, the $\ell_1$ tracking method obtains a sparse regression that can adaptively select a small number of relevant templates to optimally approximate the given test samples. 
However, it is extremely time consuming on account of solving an $\ell_1$-norm convex problem for each particle. 
For computational efficiency, Li et al. proposed real-time visual tracking with compressive sensing~\cite{Li2011Real} to solve the sparsity optimization problem by exploiting orthogonal matching pursuit (OMP). 
Recently, Zhang further improved the tracking performance by proposing a consistent low-rank sparse tracker (CLRST)~\cite{Zhang2015Robust}. 
CLRST uses temporal consistency property to prune particles adaptively and learn the candidate particles jointly with low-rank constraints. 
Compared with other approaches based on sparse linear representation and dictionary learning, the new constraints adopted by CLRST alleviate the filter degeneration problems to some extent.
Nevertheless, CLRST is unable to implement low-rank constraints globally.
However, the high-speed tracking and real-time capabilities have always been the advantages of the DCF-based trackers. 
In this paper, we learn the idea from the tradition sparse linear representation tracking framework, and use it to combine with the DCF tracking framework to make a trade-off between accuracy and efficiency.

\textbf{Analyzing Color Features for Tracking}: The choice of exploiting color features is very useful and crucial for visual tracking. 
Compared with intensity or HoG features, color features are more expert in dealing with target deformation. 
However, color features alone are not discriminative enough to separate the target from the background, and often lead to poor tracking performance when illumination is not consistent throughout a sequence.
Therefore, most successful DCF-based trackers employ the combinations of these features to inherent robustness of target deformations and color changes. 
There are two main approaches which are often adopted by DCF-based tracker, namely color names (CN)~\cite{Weijer2009Learning} and color histogram. 
CN is a linguistic color labels assigned by humans in order to represent colors in the real-world. 
It maps the RGB values to a probabilistic 11 dimensionals color representation that sums up to 1. 
Danelljan \textit{et al.} first incorporated CN into DCF tracking framework~\cite{Danelljan2014Adaptive}. 
Different from CN~\cite{Weijer2009Learning}, color histogram is based on color probability statistics. 
It followed HoughTrack~\cite{Godec2011Hough} and PixelTrack~\cite{Duffner2014PixelTrack} method to accumulate votes from each pixels. 
The area with the largest number of votes is used to estimate the position of the target. 
Bertinetto \textit{et al.} exploited color histogram as a dependent model in tracker STAPLE. 
STAPLE has shown great robustness when object target changes colors or deformation due to its excellent performance in VOT2015. 
In this paper, we investigate the influential characteristics of CN and color histogram features for visual tracking. 
Our finding indicates that the use of CN can be supportive to accurate target localization, while color histogram features can be used independently and fused at the final stage to enhance the robustness of the tracker. 
Therefore, our tracker uses two different color features simultaneously to obtain a good trade-off between accuracy and robustness. 
Moreover, we propose structural color histogram method in this paper which can enhance the reliability of color histogram features adopted by STAPLE, and make it more robust. 

\section{Tracking Formulation}\label{tracker}

\begin{figure}[tbp]
\begin{center}
   \includegraphics[width=1\linewidth]{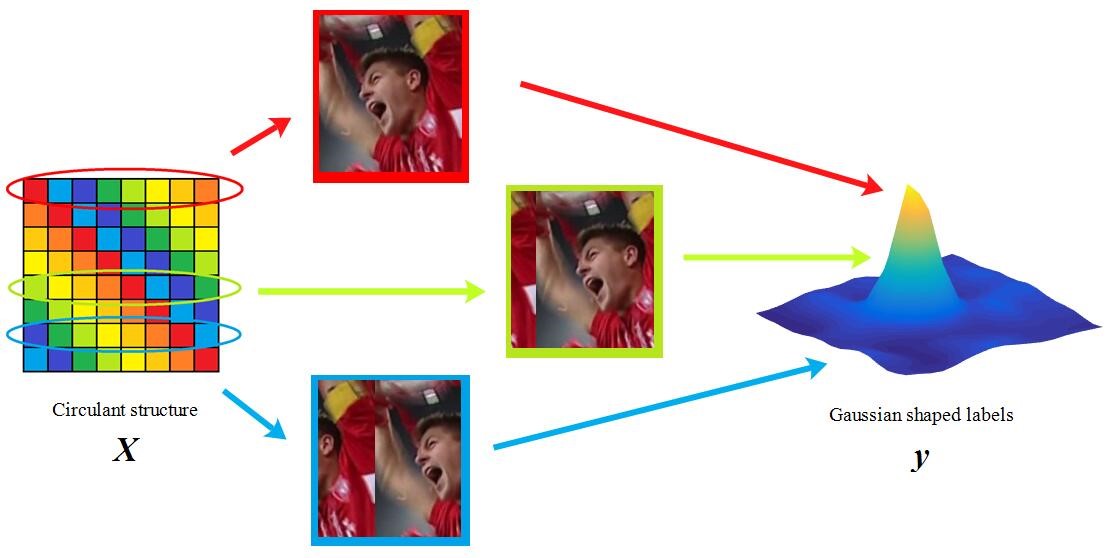}   
\end{center}
   \caption{Training process by exploiting cyclic shift to generate augmented training samples. 
   The objective is to train a classifier with both the base samples and several virtual samples (the virtual samples are obtained by circulant translating the base samples). 
   In practical applications, \textit{cosine} window is used to reduce the boundary effects during the learning stage.}\label{fig2}
\end{figure}
\begin{figure*}[!t]
\begin{center}
   \includegraphics[width=1\linewidth]{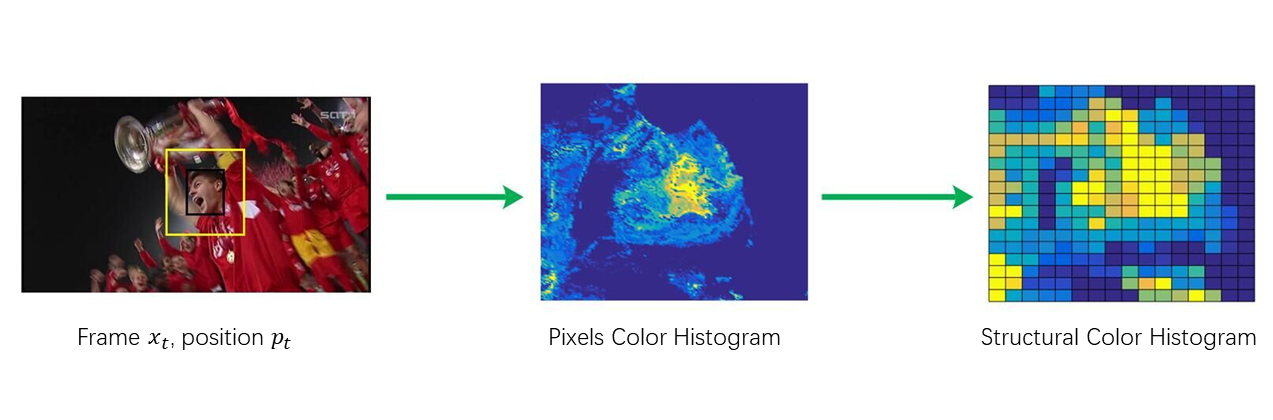}   
\end{center}
   \caption{Visualization of the structural color histogram method in frame $x_t$, the size of each block is $4\times 4$ pixels. Compared with CN and original color-based model adopted by STAPLE, the proposed structural color histogram features is more robust. The corresponding response map is independently used and fused at the final stage to enhance the complementary properties of different features}\label{fig3}
\end{figure*}

\subsection{Traditional Template Model of Discriminative Correlation Filter}
The objective of a typical tracker which is based on discriminative correlation filter is finding a function that minimizes the squared error over feature of shifted training samples ${\bm{f}_d}$ and their Gaussian shaped label $\bm{g}$~\cite{Henriques2012Exploiting}, which is shown in Fig.~\ref{fig2}. 
In Eq.~(\ref{obj1}), $D$ denotes the number of feature channels and ${\bm{h}_d}$ is the filter refers to the $d$-th channel of a vectorized frame. 
$\ast$ denotes the circular convolution operator, and $\lambda\geq 0$ is the parameter of the regularization term that controls over-fitting:
\begin{align}
\label{obj1}
\arg\underset{\bm{\mathit{h}}}{\min}
&
\sum\limits_{d=1}^{D}\left\|
\bm{f}_d\ast{\bm{h}}_d-\bm{g}\right\|_2^2+\lambda\sum\limits_{d=1}^{D}\left\|\bm{h}_d\right\|_2^2.
\end{align}
In order to achieve computational efficiency, DCF trackers are typically learned in the frequency domain~\cite{Kumar2005Correlation}. 
Hence, Eq.~(\ref{obj1}) can be written as:
\begin{align}
\label{obj2}
\arg\underset{\hat{\bm{h}}}{\min}
&
\sum\limits_{d=1}^{D}\left\|
\hat{\bm{h}}_d^H \textbf{diag}{(\hat{\bm{f}}}_d)-\hat{\bm{g}}\right\|_2^2+\lambda\sum\limits_{d=1}^{D}\left\|\hat{\bm{h}}_d\right\|_2^2.
\end{align}
where the hat, $\hat{}$, denotes the Discrete Fourier Transform (DFT).
Minimizing the error of Eq.~(\ref{obj2}) has a closed-form solution, which is:
\begin{equation}
\label{eq2}
 \bm{h}_d=\frac{\overline{\bm{g}}\bm{f}_d}{\sum_{d=1}^{D}\overline{{\bm{f}_d}}{\bm{f}_d}+\lambda}.
\end{equation}

As the target appearance may vary during tracking, the filter $\bm{h}_d$  need to be updated with the corresponding appearance in a new frame. 
The filters obtained in the previous frames are linearly summed as the result of history template model. 
In spite of its simplicity, this closed-form solution does not take the  historical information into account and it suffers from the boundary inevitably. 
To address these issues, we propose a low-rank correlation filter by using $\ell_2$-norm. 
As for scale-detection, our approach follows the DSST~\cite{danelljan2014accurate} tracking framework.

\subsection{Learning template of Low-Rank Discriminative Correlation Filter}
To improve the robustness of filters, we add two constraints into the objective function. 
First, in order to take the model template history information into account, our approach tends to make the filter learn from the current frame as similar as the filter obtained in the previous frame. 
Therefore, we introduce a new constraint to the objective function, which is:
 \begin{equation}
\label{eq4}
\min\lambda_2\left\|\bm{h}-\bm{h}^{t-1}\right\|_2^2,
\end{equation}
where $t-1$ denotes the previous frame index and we omit the index in the current frame for simplification, $\bm{h}=\bm{h}^t$.
The vectorized filters obtained by Eq.~(\ref{eq4}) tend to be low-rank, which allows them to be summed in a low dimensional space. 
We will demonstrate this property in the Subsection~\ref{lr}. 

Second, as there is no previous learned filter in the first frame, and in order to make the filter that obtained in the first frame to be discriminative, our approach adopts an extensive prior assumption for the filter in the first frame, which is:
\begin{equation}
\label{eq5}
\bm{h}_m=\bm{m}\odot \bm{h}
\end{equation}
where $\bm{h_m}$ denotes the learned filter with mask. 
We define the mask as a binary matrix which sets the value of target center area as $1$, and restricts the background area as $0$ during the feature extraction~\cite{Galoogahi2017Learning}. 
On the other hand, the mask can effectively suppress the boundary effects and expand the search region, which enlarges the training sample sets.
However, introducing Eq.~(\ref{eq4}) and (Eq.~\ref{eq5}) prohibits a closed-form solution of Eq.~(\ref{obj2}). 
We employ the Augmented Lagrangian Method (ALM) and a dual variable $\bm{h}_c$ for the optimization, with the constraint ${\bm{h}_c}={\bm{m}\odot\bm{h}}$. 
Therefore, the proposed objective function by our approach can be rewritten as:
 \begin{equation}
\label{eq6}
\begin{aligned}
&\mathcal{L}\left(\bm{\mathit{\hat{h}_c}},\bm{h},\bm{\mathcal{H}}|\bm{m}\right)=\left\|
\hat{\bm{h}}_c^H \textbf{diag}{(\hat{\bm{f}}}_d)-
\hat{\bm{g}}\right\|_2^2+\frac{\lambda_1}{2}\left\|\hat{\bm{h}}_m\right\|_2^2\\&
+[\bm{\mathcal{H}}(\bm{h_c}-\bm{h}_m+\overline{\bm{\mathcal{H}}(\bm{h}_c-\bm{h}_m)}]+\mu\|\bm{h}_c-\bm{h}_m\|_2^2\\
&+\frac{\lambda_2}{2}\|\bm{h}_c-\bm{h}^{t-1}\|_2^2
\end{aligned}
\end{equation}
Where $\bm{\mathcal{H}}$ is the Lagrange multiplier, $\lambda_1$ and $\lambda_2$ are the regularization terms for $\bm{h}_m$ and $\bm{h}_c$ respectively. The augmented Lagrangian Eq.~(\ref{eq6}) can be iteratively optimized by ALM, which alternately solves the following two sub-problems at each iteration:
\begin{equation}
\label{eq7}
\left\{\begin{aligned}
&\hat{\bm{h}}_c^{i+1} =\arg\underset{\bm{h}_c}{\min}~\mathcal{L}\left(\bm{\hat{h}}_c,\bm{h}^i,\bm{\mathcal{H}}^i|\bm{m}\right)\\
&\hat{\bm{h}}_m^{i+1}=\arg\underset{\bm{h}}{\min}~\mathcal{L}\left(\bm{\hat{h}}_c^{i+1},\bm{h},\bm{\mathcal{H}}^i|\bm{m}\right)\\
\end{aligned}\right. 
\end{equation}

And the Lagrange multiplier can be updated as:
\begin{equation}
\label{eq8}
\bm{\mathcal{H}^{i+1}} = \bm{\mathcal{H}^i} +\mu\left(\bm{\hat{h}}_c^{i+1}-\bm{\hat{h}}^{i+1}\right),
\end{equation}
where $\mu$ denotes the constraint penalty parameter, which can be updated as:
\begin{equation}
\mu^{i+1} =\beta\mu^i.
\label{eq9}
\end{equation}                                                                
The minimizations of Eq.~(\ref{eq7}) have an approximate solution:
\begin{equation}
\label{eq10}
\left\{\begin{aligned}
&\hat{\bm{h}}_c^{i+1} =\frac{\hat{\bm{f}}\odot\hat{\bm{g}}^H-\bm{\mathcal{H}}+\mu\hat{\bm{h}}_m+\lambda_2\hat{\bm{h}}^{t-1}}{\hat{\bm{f}}\odot\hat{\bm{f}}^H+(\mu+\lambda_2)\bm{I}}\\
&\bm{h}_m^{i+1}=\bm{m}\odot\mathscr F^{-1}{\frac{\bm{\mathcal{H}+2\mu\bm{h}}_c^{i+1}}{(2\mu+\lambda_1)}}\\
\end{aligned}\right. 
\end{equation}

Notice the Eq.~(\ref{eq10}) is computed in the frequency domain and in the original domain iteratively, requiring the Inverse Discrete Fourier Transform in each iteration. 
Similar to most DCF-based trackers, our approach follows the online update strategy of the model templates. 
The filter $\bm{h}$ in a new frame $t$ can be formulated as:
\begin{equation}
\bm{h}_{\textrm{model}}^t =(1-\eta_1)\bm{h}_{\textrm{model}}^{t-1}+\eta_1\bm{h}
\label{eq11}
\end{equation}
where $\bm{h}_{\textrm{model}}^{t-1}$ denotes DCF template model in frame $t-1$ , $\bm{h}$ denotes the learned template model of the current frame, $\eta$ is the parameter of learning rate.
To locate the target in a new frame, a patch $\bm{z}$ needs to be extracted at the predicted target location. The target’s new position is found by maximizing the response map.
\begin{eqnarray}\label{eq12}
  	\bm{r}(\bm{z})=\mathscr F^{-1}\left({\bm{\hat{h}}}_{\textrm{model}}^{t}\odot{\bm{\hat z}}_{t}\right)
\end{eqnarray}
Where denotes $\mathscr F^{-1}$ the inverse of DFT, and $\odot$ denotes point-wise multiplication. We adopt the interpolation method in~\cite{Danelljan2016Convolutional} to maximize response map per each correlation output.

\begin{algorithm}[t]
\begin{algorithmic}
\vspace{0.03in}
\STATE
\textbf{Require:}
Extracted image patch features $\bm{f}$, desire correlation output $\bm{g}$, binary mask $\bm{m}$;

\textbf{Ensure:}
Optimized filter $\bm{\hat{h}}$;

\textbf{Procedure:}

\STATE 1: Initialize filter $\hat{\bm{h}}^0$ by ${\bm{h}^{t-1}}$;

\STATE 2: Initialized Lagrange multiplier: $\bm{\mathcal{H}^0}$ ${\leftarrow}$ $\textbf{0}$;

\STATE 3: repeat

\STATE 4: Compute $\hat{\bm{h}}_c^{i+1}$ and $\hat{\bm{h}}_m^{i+1}$ using Eq.~(\ref{eq10});

\STATE 5:Update the Lagrange multiplier $\bm{\mathcal{H}^{i+1}}$ and penalty $\mu$ using Eq.~(\ref{eq8}) and Eq.~(\ref{eq9});

\STATE 6:until stop condition
\vspace{0.03in}
\end{algorithmic}
\caption{Low-rank Constrained Filter optimization.}
\label{alg1}
\end{algorithm}\subsection{Learning Template of Color Histogram Score}
Our tracker uses color histogram features as an independent color-based model to support further accuracy. 
This strategy was first adopted by STAPLE~\cite{Bertinetto2016Staple} tracker, which the color histogram features are considered as quantised RGB colors.
Comparing to HOG, color histogram features are more robust to object color changes, deformations and partial occlusion. In a frame $t$, the color histograms of foreground and background can be extracted according to the target position.
Tracker STAPLE obtains histogram score by accumulating votes from each pixel, and the color histogram score is considered as average votes. 
Let $F$ denote object region, and $B$ denote background region.
The votes of pixel can be computed as:
\begin{eqnarray}\label{eq13}
     p^j(F)={hist{\_F}}(b(u)_i)\\
     p^j(B)={hist{\_B}}(b(u)_i)\\
  	\beta_{t}^{j}=(\frac{(p^j(F))}{(p^j(F)+p^j(B)+\lambda_3)}
\end{eqnarray}
where $p^j(F)$ and $p^j(B)$ are the probabilities of object region and background region respectively. $b(u)_i$ is the pixel index of the color histograms. The purpose of the parameter $\lambda_3$ is to prevent the denominator from being zero during the calculation, and $j$ denotes the feature dimensions. 
The obtained vote $\beta_{t}^{j}$ is the probability of each pixel that belong to the object target. 
After integrating all these votes for each pixel, the area with the largest probability is considered as the center of the target. 
STAPLE gives a score to the template color histogram according to the obtained probability. 
The score can be used later to fuse with the response of DCF model template.
We evaluate STAPLE tracker's performance on OTB100~\cite{Wu2015Object} and TC128~\cite{Liang2015Encoding} benchmarks. 
Our finding suggests that the independent color-based model is able to inherent robustness to either color changes or deformation. 
However, when the colors of foreground and background are highly similar, this color probability model may overestimate the confidence of target's position and lead to poor tracking performance. 
Therefore, our approach designed a novel structural color histogram features as an independent color-based model to enhance the robustness, while the HOG and CN features emphasize accurate target localization. 
Our approach divides original image into $4\times 4$ pixels image blocks. 
We choose the pixel with largest probability to represent the entire image block. 
In order to make the color histogram features more discriminatory, we sort all the image blocks according to probability value and discard $20\%$ less significant blocks with small value. 
This strategy makes the color histogram features structural and also effectively increase the reliability of color histogram features.
Our approach followed the color statistic model which is adopted in tracker STAPLE~\cite{Bertinetto2016Staple}, but our method accumulates votes from each block instead of pixel. 
The image-block method and structural color histogram features are shown in Fig.~\ref{fig3}.

The model parameters are updated as:
\begin{eqnarray}
  	p_t(F)=(1-\eta_{\textrm{hist}})p_{t-1}(F)+\eta_{\textrm{hist}}p_t^{'}(F),\label{eq14} \\
    p_t(B)=(1-\eta_{\textrm{hist}})p_{t-1}(B)+\eta_{\textrm{hist}}p_t^{'}(B).\label{eq15}
\end{eqnarray}
We followed a linear combination of template and histogram scores which is frequently used in classical DCF trackers, denotes balancing coefficient factor:
\begin{eqnarray}\label{eq16}
  	\bm{r}(\bm{z})=(1-\alpha)\bm{r}_{\textrm{tmpl}}(\bm{z})+\alpha \bm{r}_{\textrm{hist}}(\bm{z})
\end{eqnarray}

\begin{algorithm}[t]
\begin{algorithmic}
\vspace{0.03in}
\STATE
\textbf{Input:}
Initial target bounding box $\bm{x_0}$, initial target position $\bm{p_0}$;

\textbf{Repeat:}

\STATE 1: Extract HOG and CN features in searching window of frame $t$, according to target position in $t-1$; 

\STATE 2: Compute the DCF response map $\bm{r}(\bm{z})$ using Eq.~(\ref{eq12});
\STATE 3: Compute Structural Color Histogram response map using Eq.~(\ref{eq13});
\STATE 4: Compute the final response map $\bm{r(z)}$ in frame $t$ using 
Eq.~(\ref{eq16});
\STATE 5: Find the possible target position $\bm{p}_t$ by $\bm{r(z)}$;
\STATE 6: Estimate target scale according to $\bm{p}_t$;
\STATE 7: Update model using Eq.~(\ref{eq11}), Eq.~(\ref{eq14}) and Eq.~(\ref{eq15});

\STATE \textbf{end}
\STATE \textbf{Until} End of video sequences;
\STATE \textbf{Output:} Target position $\bm{p}_t$.
\vspace{0.03in}
\end{algorithmic}
\caption{ILRCSCH tracking algorithm.}
\label{alg2}
\end{algorithm}

\section{Performance Evaluation}\label{experiment}

This section introduces comprehensive experimental results of the proposed ILRCSCH tracker. 
Evaluation methodology are discussed in Subsection~\ref{111}, Implementation details are given in Subsection~\ref{222}. 
The performance of ILRCSCH tracker and comparison results with state-of-the-art methods on four recent standard benchmarks are reported in Subsection~\ref{333}, Subsection~\ref{444} and Subsection~\ref{555}. 
In Subsection~\ref{lr}, we make rank statistic on our tracker and other nine state-of-art trackers to prove our tracker has low-rank characteristics. 

\subsection{Evaluation Methodology }\label{111}
We evaluate the proposed method on OTB-2013~\cite{Wu2013Online}, OTB100~\cite{Wu2015Object}, TC128~\cite{Liang2015Encoding} and VOT2016~\cite{Kristan2016The} benchmarks. 
For OTB datasets, we use mean overlap precision (OP) and area-under-the-curve (AUC) to rank the trackers under OPE (one pass evaluation). 
The overlap precision is computed as the percentage of bounding box in a video sequence whose intersection-over-union (IoU) score is larger than a certain threshold. 
The mean OP of all video sequences is plotted over the range of IoU thresholds to obtain the success plot. 
We use area-under-the-curve (AUC) of success plots to rank the trackers. For VOT2016, tracking performance is evaluated in terms of expect average overlap rate (EAO) and robustness (failure rate). 
In VOT2016, trackers will be restarted in the case of tracking failure (when the overlap rate is lower than a threshold). 
The readers are encouraged to read~\cite{Wu2013Online, Wu2015Object} and~\cite{Kristan2016The} for more details.

\begin{table}[!t]
\footnotesize
\renewcommand{\arraystretch}{1.3}
\caption{Parameters setting in our experiments.
\label{setting}}{
\begin{tabular}{lc}
\hline
Learning rate (template of HOG and CN) $\eta_{\textrm{tmpl}}$&0.95\\
Learning rate (histogram) $\eta_{\textrm{hist}}$&0.04\\
\#bins color histograms & $32\times 32\times 32$\\
CN cell size&$4\times 4$\\
HOG cell size&$4\times 4$\\
Scale-step&1.01\\
Search area scale&4\\
Pixels of each block&$4\times 4$\\
Balancing coefficient factor $\alpha$&0.1\\
\hline
\end{tabular}}{}
\end{table}

\subsection{Implementation Details}\label{222}
Our tracker is implemented in Matlab code on a laptop equipped with an Intel Core i7-7700HQ @2.48GHz and 8GB memory. 
We employ three different handcrafted features, including HOG, Color Names (CN) and proposed structural color histogram features. 
The combination of HOG and CN features are used in DCF tracking to emphasize accurate target localization, while the structural color histogram features work as an independent color-based model and fuse at the final stage to enhance the robustness of the tracker. 
Similar to recent DCF-based trackers, we adopt fhog from Piotr's Computer Vision Matlab Toolbox as HOG features multiplied by a Hann window~\cite{Bolme2010Visual} in the implementation of the source code. 
We give the parameters value of the paper In Tab.~\ref{setting}, which may helpful to the readers. 
Note the parameters of our tracker were kept constant throughout all experiments and do not require fine-tuning.

\begin{figure*}[!t]
\begin{center}
   \includegraphics[width=0.24\linewidth]{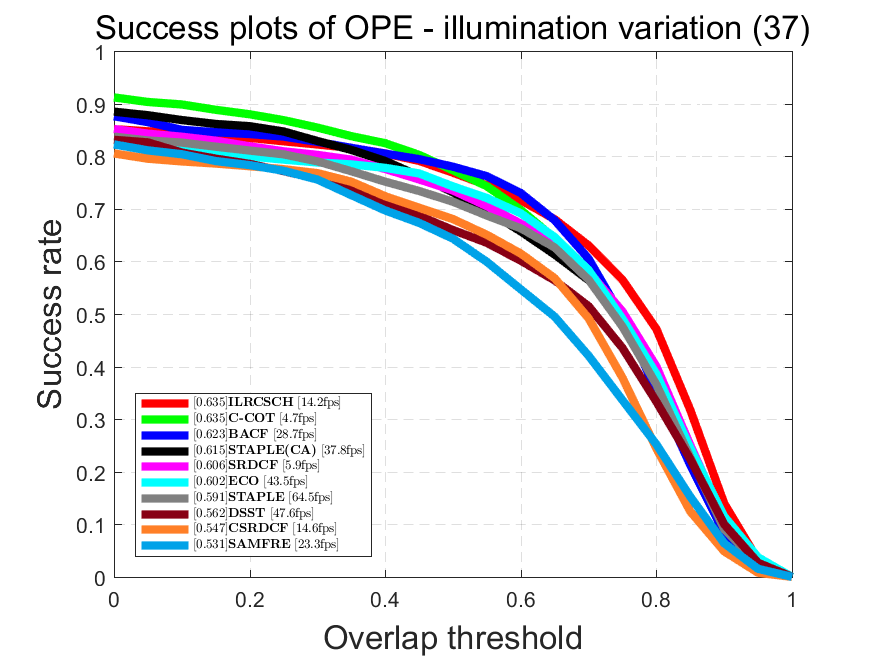}   
   \includegraphics[width=0.24\linewidth]{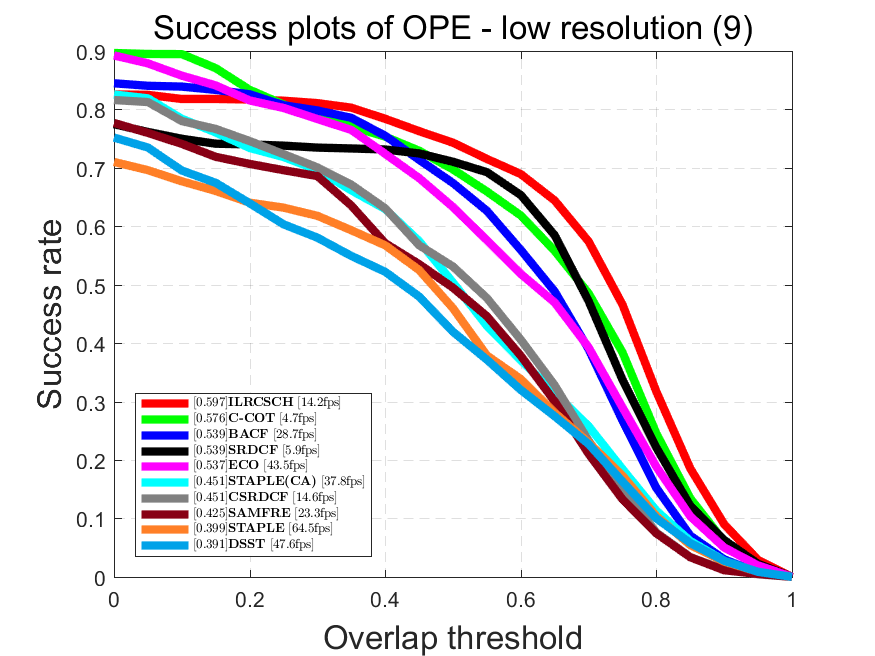}
   \includegraphics[width=0.24\linewidth]{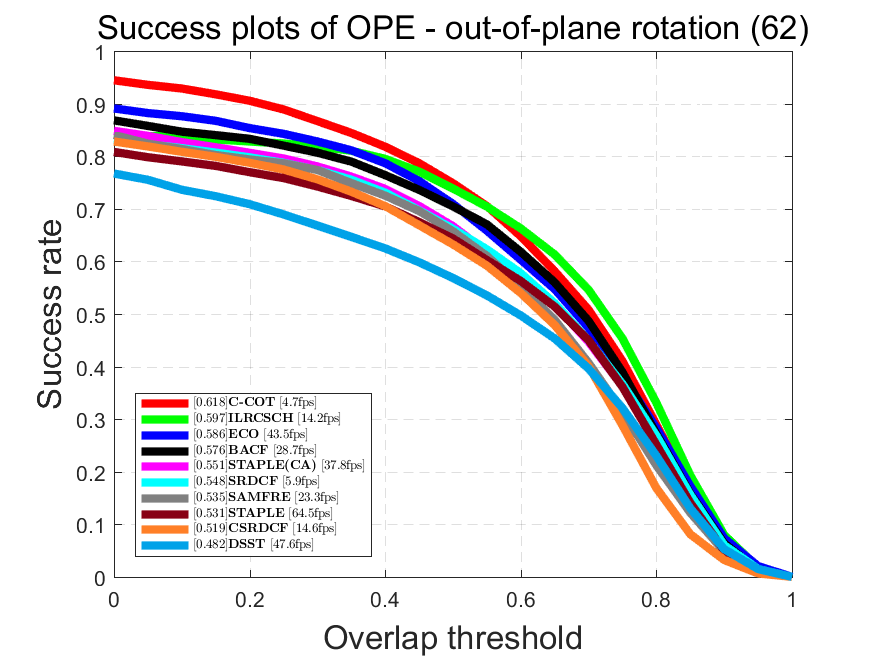}   
   \includegraphics[width=0.24\linewidth]{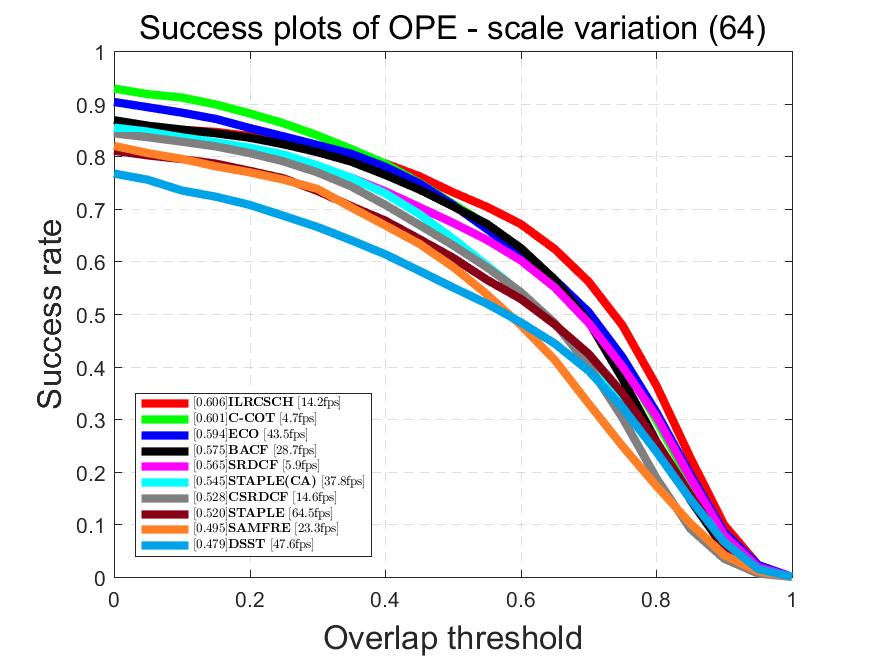}  
\end{center}
   \caption{OTB-100 dataset attribute-based analysis. The results demonstrate the superior performance of our ILRCSCH compared with other methods.}\label{fig5}
\end{figure*}

\begin{figure}[!]
\begin{center}
   \includegraphics[width=0.48\linewidth]{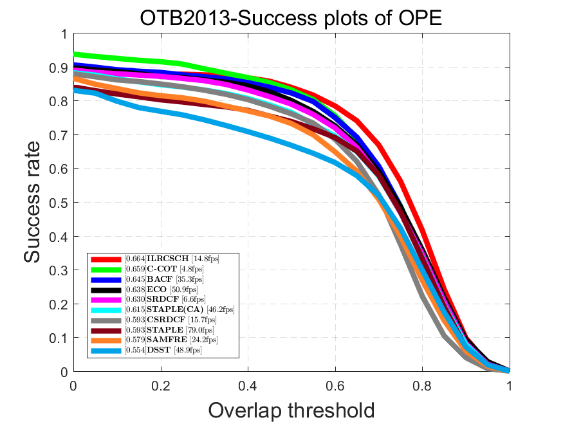}   
   \includegraphics[width=0.48\linewidth]{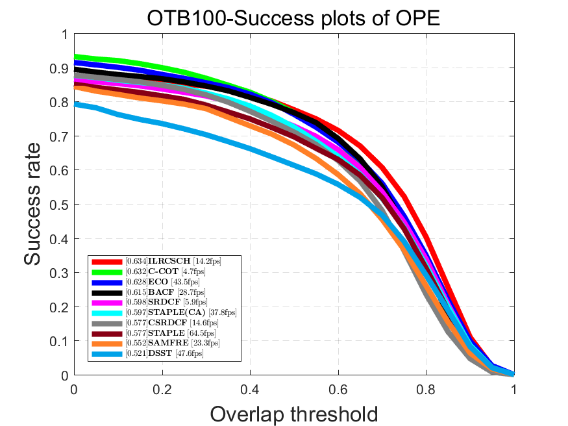}   
\end{center}
   \caption{Success plots on OTB2013 and OTB2015 benchmark comparisons with the state-of-the-art hand-crafted based trackers. AUCs and FPSs are reported in the legends. The proposed ILRCSCH tracker is ranked the top.}\label{fig4}
\end{figure}

\subsection{Comparison results on OTB2013 and OTB100 benchmark}\label{333}
The OTB2013\cite{Wu2013Online} and OTB100~\cite{Wu2015Object} benchmarks contain 50 and 100 video sequences respectively. 
The evaluation of the tracker performance is measured by precision and success plots. 
We test our proposed tracker ILRCSCH with 16 state-of-the-art methods, including ECO~\cite{Danelljan2016ECO}, C-COT~\cite{Danelljan2016Beyond}, BACF~\cite{Galoogahi2017Learning}, CSRDCF~\cite{Lukezic2017Discriminative}, STAPLE(CA)~\cite{mueller2017context}, STAPLE~\cite{Bertinetto2016Staple}, SRDCF~\cite{Danelljan2015Learning}, LMCF~\cite{Wang2017Large}, LCT~\cite{Ma2015Long}, DSST~\cite{danelljan2014accurate}, SAMF~\cite{Li2014A}, CN~\cite{Danelljan2014Adaptive}, KCF~\cite{Henriques2015High}, CSK~\cite{Henriques2012Exploiting}, Struck~\cite{Hare2016Struck} and TLD~\cite{Kalal2012Tracking}. 
To reduce clutter in the graphs, we only present the results of those recent top-performing baselines. 
In Tab.~\ref{otb50} and Tab.~\ref{otb100}, we present the experimental results of success rate (AUC), overlap precision (OP) and frames per second (FPS) under OTB2013 and OTB100 benchmarks respectively. 
The pipeline graphs are displayed in Fig.~\ref{fig4}. 
The trackers' ranking order prioritize the value of AUC.

\begin{table*}[!t]
\centering
\caption{\label{OTB2013}  The tracking results of trackers based on hand-crafted features on OTB2013 benchmark. The top three results are shown in {\color{red}{red}}, {\color{blue}{blue}} and {\color{brown}{brown}}}
\resizebox{6.8in}{!}{
\begin{tabular}{ccccccccccc}
\hline
&ILRCSCH & C-COT & BACF & ECO & SRDCF & STAPLE(CA) & CSRDCF & STAPLE & SAMF & DSST\\
\hline 
AUC &  {\color{red}{0.664}} & {\color{blue}{0.659}} &  {\color{brown}{0.645}} & 0.638 & 0.630 & 0.615 & 0.593 & 0.593 & 0.579 & 0.554\\
OP & {\color{brown}{0.852}} & {\color{red}{0.875}} & 0.849 &  {\color{blue}{0.856}} & 0.838 & 0.832 & 0.825 & 0.782 & 0.785 & 0.737\\
\hline 
Mean FPS ($\%$) on CPU & 14.8 & 4.8 & 35.3 & {\color{blue}{50.9}} & 6.6 & 46.2 & 15.7 & {\color{red}{79.0}} & 23.3 & {\color{brown}{47.6}}\\
\hline
\end{tabular}}
\label{otb50}
\end{table*}

\begin{table*}[!t]
\centering
\caption{\label{OTB2015}  The tracking results of trackers based on hand-crafted features on OTB100 benchmark. The top three results are shown in {\color{red}{red}}, {\color{blue}{blue}} and {\color{brown}{brown}}}
\resizebox{6.8in}{!}{
\begin{tabular}{ccccccccccc}
\hline
  & ILRCSCH & C-COT & ECO & BACF & SRDCF & STAPLE(CA) & CSRDCF & STAPLE & SAMF & DSST\\
\hline
AUC & {\color{red}{0.634}} & {\color{blue}{ 0.632}} & {\color{brown}{0.628}} & 0.615 & 0.598 & 0.597 & 0.577 & 0.577 & 0.552 & 0.521\\
OP    & {\color{brown}{0.833}} & {\color{red}{0.855}} & {\color{blue}{0.837}} & 0.816 & 0.808 & 0.801 & 0.786 & 0.782 & 0.751 & 0.695\\
\hline
 Mean FPS ($\%$) on CPU & 14.6 & 4.7 & {\color{blue}{50.9}} & 28.7 & 5.9 & 37.8 & 14.6 & {\color{red}{64.5}} & 24.2 & {\color{brown}{48.9}}\\
\hline
\end{tabular}}
\label{otb100}
\end{table*}

Tab.~\ref{otb50} and Tab.~\ref{otb100} compare the proposed tracker (ILRCSCH) with the state-of-art handcrafted feature-based DCF trackers on the OTB2013 and OTB100 two different benchmarks, where our tracker achieved the best performance in terms of AUC. 
More particularly, ILRCSCH achieved the best AUC (0.664) on OTB2013 followed by C-COT (0.659), and ECO (0.638). 
On OTB100, ILRCSCH (0.645) outperformed C-COT (0.632) and ECO (0.638). These results fully illustrate the importance of utilizing low-rank constraints to learn more robust filters from handcrafted features. 
This evaluation also shows that our strategy is more effective than CSRDCF tracker, which also uses mask for the learned filters. 
This is mainly because, unlike ILRCSCH, the mask adopted by CSRDCF is obtained from color histogram features, and the mask is required to be used for all learned filters to reduce boundary effects. 
However, the color information is not always robust and reliable throughout the whole tracking process, which may lead to poor tracking performance for the trackers. 
Different from CSRDCF, our approach only requires mask for the learned filter at the first frame as a low-rank prior assumption to make our model template as similar as the object target. 
In the later stage, we adaptively obtain the history information of the template through low-rank constraints (Eq.~(\ref{eq4})). 
This adaptive learning strategy is able to make our tracker more robust and accurate. 
Tab.~\ref{otb50} and Tab.~\ref{otb100} also report the tracking speed of all trackers. The best tracking speed belongs to STAPLE (64.5 FPS) followed by ECO (50.9 FPS) and DSST (47.6 FPS). 
Except ECO, such higher speed trackers came at the cost of much lower accuracy compared to ILRCSCH. 
Our tracker obtained the speed of 14.6 FPS which is almost 3 times faster than C-COT.

\begin{table*}[!t]
\centering
\caption{\label{otbophc}  The tracking results of trackers based on hand-crafted features on TC128 benchmark. The top three results are shown in {\color{red}{red}}, {\color{blue}{blue}} and {\color{brown}{brown}}}
\resizebox{6.8in}{!}{
\begin{tabular}{ccccccccccc}
\hline
 & ECO & C-COT & ILRCSCH & STAPLE(CA) & STAPLE & BACF & SRDCF & CSRDCF & SAMF & DSST\\
\hline
AUC & {\color{red}{0.543}} & {\color{blue}{ 0.529}} & {\color{brown}{0.527}} & 0.504 & 0.496 & 0.485 & 0.484 & 0.474 & 0.462 & 0.408\\
OP    & {\color{red}{0.731}} & {\color{blue}{0.729}} & {\color{brown}{0.706}} & 0.680 & 0.676 & 0.667 & 0.663 & 0.646 & 0.628 & 0.546\\
\hline
 Mean FPS ($\%$) on CPU & {\color{brown}{46.2}} & 4.9 & 10.9 & 41.5 & {\color{red}{66.1}} & 32.4 & 7.3 & 15.2 & 16.9 & {\color{blue}{52.8}}\\
\hline
\end{tabular}}
\label{otbophc}
\end{table*}

\subsubsection{Attribute based analysis on OTB-2015}
To further evaluate the performance of our method, we give the attribute-based analysis of our tracker on the OTB-100 dataset which contains 100 video sequences in Fig.~\ref{fig5}. 
These attributes are useful to describe the different challenges in the tracking problems. 
The attributes are shown by the success plots. 
Each plot title includes the number of videos is related to the respective attributes. 
To reduce clutter in the graphs, only 10 trackers and 4 attributes graphs are displayed.
Fig.~\ref{fig5} demonstrates our tracker performs well against the other trackers. 
Especially when object target suffers scale variation, illumination variation and out-of-plane rotation, our method is still very robust.

\begin{table}[!t]
\footnotesize
\centering
\renewcommand{\arraystretch}{1.3}
\caption{\label{feature}  The results in the table illustrate the impact of using different hand-crafted features on the tracking performance of the proposed ILRCSCH tracker. The combination of three hand-crafted features achieves the best tracking results.}
{
\begin{tabular}{lccc}
\hline
ILRCSCH&HOG&HOG+CN&HOG+CN+HIST(structured)\\
\hline
AUC&0.495&0.523&0.527\\
OP&0.659&0.704&0.706\\
FPS&27.4&10.1&10.0\\
\hline
\end{tabular}}
\label{feature}
\end{table}

\subsection{Comparison results on TC128}\label{444}
The TC128 benchmark contains 128 color video sequences. 
We first analyze the effect of two different color features (CN and proposed Structural Color Histogram) on ILRCSCH tracker. 
The tracking results are given in Tab.~\ref{feature}.

The data in Tab.~\ref{feature} well demonstrates the color features can provide complementary support to the final response map due to the object's appearance can vary significantly during the tracking process. 
It is well known that template model (HOG) depend on the spatial configuration of the object and usually perform poorly when this changes quickly. 
The proposed ILRCSCH tracker can rely on the strengths of the DCF templates (HOG and CN) and an independent color-based model (proposed Structural Color Histogram). 
This strategy allows our tracker inherit accuracy and robustness to both color changes and deformations.

We also made comparisons with the state-of-the-art handcrafted feature-based trackers on TC128 benchmarks. 
The results are given in Tab.~\ref{otbophc}.

\begin{table}[!t]
\footnotesize
\centering
\renewcommand{\arraystretch}{1.3}
\caption{\label{vot}  The VOT2016 results of the proposed ILRCSCH tracker, the EAO and Failures show that our hand-crafted feature-based well against the state-of-the-art deep feature-based trackers.}
{
\begin{tabular}{lccccc}
\hline
Tracker&ILRCSCH&TCNN&EBT&MDNet\_N&DeepSRDCF\\
\hline
EAO&0.27&0.32&0.29&0.26&0.28\\
Failures&16&8&11&18&17\\
Overlap&0.58&0.58&0.52&0.56&0.57\\
AUC&0.41&0.49&0.37&0.46&0.43\\
\hline
\end{tabular}}
\label{vot}
\end{table}

\subsection{Comparison with deep Feature-based trackers on VOT2016}\label{555}
In this part, we compare the proposed ILRCSCH tracker with several deep feature-based tracker on VOT2016 benchmark. 
VOT2016 dataset includes 60 video sequences from VOT2015 with improved annotations. 
The benchmark evaluated over 70 trackers such as C-COT~\cite{Danelljan2016Beyond}, STAPLE~\cite{Bertinetto2016Staple}, MLDF~\cite{Wang2016Visual}, TCNN~\cite{Kang2016Object}, SSAT ~\cite{Nam2015Learning} and so forth. 
In Table 6 we present the comparison of the proposed ILRCSCH tracker with part of the state-of-art deep feature-based trackers, which contains TCNN, EBT~\cite{Zhu2016Beyond}, MDNet\_N~\cite{Nam2015Learning}, DeepSRDCF~\cite{Danelljan2016Convolutional}. 
The results demonstrate the proposed ILRCSCH tracker well against those trackers which are applied computationally intensive deep features, even though we simply use handcrafted features. 

\begin{figure}[tbp]
\begin{center}
   \includegraphics[width=1\linewidth]{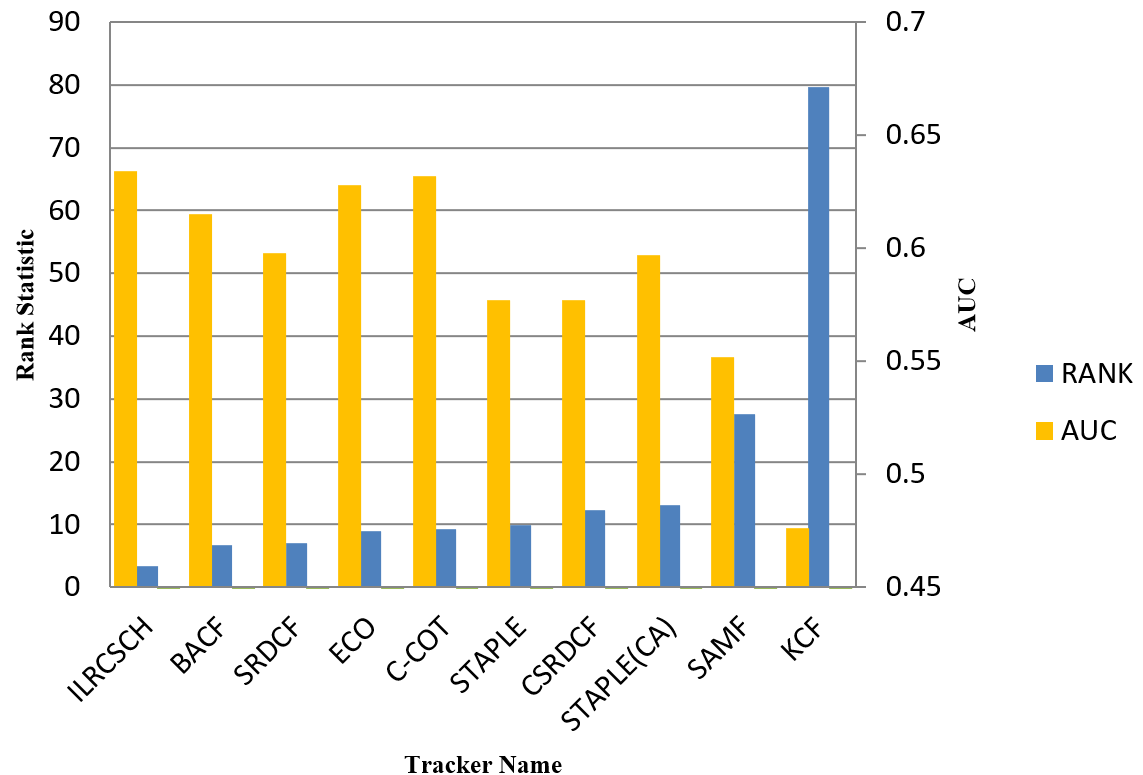} 
\end{center}
   \caption{Rank statistic and AUC scores of the state-of-the-art trackers' filters on OTB2015 benchmark. The proposed ILRCSCH tracker achieves the lowest rank (3.40) and highest AUC score (0.634).}\label{fig6}
\end{figure}
\subsection{Rank Statistic Experiment}\label{lr}
Experimental results for rank statistic of ten state-of-art-trackers on OTB100 are given in Fig.~\ref{fig6}. 
We save the ten trackers' filter results that obtained from each frame for all video sequences (100 video sequences totally) on OTB100 dataset separately. 
Then vectorize all the filters and group them into a filter matrix according to their video sequences, then calculate the rank of each matrix. 
Dividing the value of the ranks by the number of frames that corresponding to their video sequences, and summing up these data for one hundred video sequences. 
The final rank statistic results are displayed in blue histograms in Fig.~\ref{fig6}. 
The AUC plots are also shown in orange histograms. Fig.~\ref{fig6} well illustrates those top performance trackers tend to be low rank (like ECO and C-COT). 
Our approach uses the most direct  $ \ell_2$-norm low-rank constraint in the objective function, so that the filter of the current frame is as similar as possible to the filter of the previous frame. 
This strategy allows the filter of the current frame effectively inherit the history information from the previous frame filter. 
Therefore, the proposed ILRCSCH tracker achieves the best tracker results in terms of AUC plots on OTB100 benchmark.

\section{Conclusion}\label{conclusion}

In this paper, we proposed ILRCSCH tracker which is based on three handcrafted features. 
By exploiting the $ \ell_2$-norm and the binary mask constraints, the learned filter can effectively obtain a lot of historical information from the previous frames and suppress the boundary effects due to the circulant assumption. 
The two different color features also provide complementary support for locating the target position. 
The CN is used for accurate target localization, while the designed structural color histogram features work as an independent color-based model and fused at the final stage to enhance the robustness of the tracker. 
Given its efficiency and simplicity, the proposed ILRCSCH tracker is a suitable choice for real-time applications. 
Furthermore, our tracker is able to remain robustness in challenging situations like illumination variations, out-of-plane rotations and partial occlusion.

\medskip


\end{document}